\documentclass[a4paper,twoside]{article}
\usepackage{epsfig}
\usepackage{natbib}
\usepackage{subcaption}
\usepackage{calc}
\usepackage{amssymb}
\usepackage{amstext}
\usepackage{amsmath}
\usepackage{amsthm}
\usepackage{multicol}
\usepackage{pslatex}
\usepackage{plain}
\usepackage{SCITEPRESS}     

\begin{document}

\title{An integrated recurrent neural network and regression model with spatial and climatic couplings for vector-borne disease dynamics}

\author{\authorname{Zhijian Li \sup{1}, Jack Xin\sup{1} and Guofa Zhou\sup{2}}
\affiliation{\sup{1} Department of Mathematics, UC Irvine, Irvine, CA 92617, USA}
\affiliation{\sup{2}Program in Public Health, School of Medicine, UC Irvine, Irvine, CA 92617, USA.}
\email{\{zhijil2,jack.xin, zhoug\}@uci.edu}
}

\keywords{Geospatial and climatic data, integrated spatio-temporal network model, vector-borne disease forecasting .}


\abstract{We developed an integrated recurrent neural network and nonlinear regression spatio-temporal model for vector-borne disease evolution. 
We take into account climate data and seasonality as external factors that correlate with disease transmitting insects (e.g. flies), also spill-over infections from neighboring regions surrounding a region of interest. The climate data is encoded to the model through a quadratic embedding scheme motivated by recommendation systems. The neighboring regions' influence is modeled by a long short-term memory neural network. The integrated model is trained by stochastic gradient descent and tested on leishmaniasis data in Sri Lanka from 2013-2018 where infection outbreaks occurred. Our model out-performed ARIMA models across a number of regions with high infections, and an associated ablation study renders support to our modeling hypothesis and ideas.}

\onecolumn \maketitle \normalsize \setcounter{footnote}{0} \vfill
\begin{figure*}[ht!]
    \centering
    \includegraphics[scale=0.14]{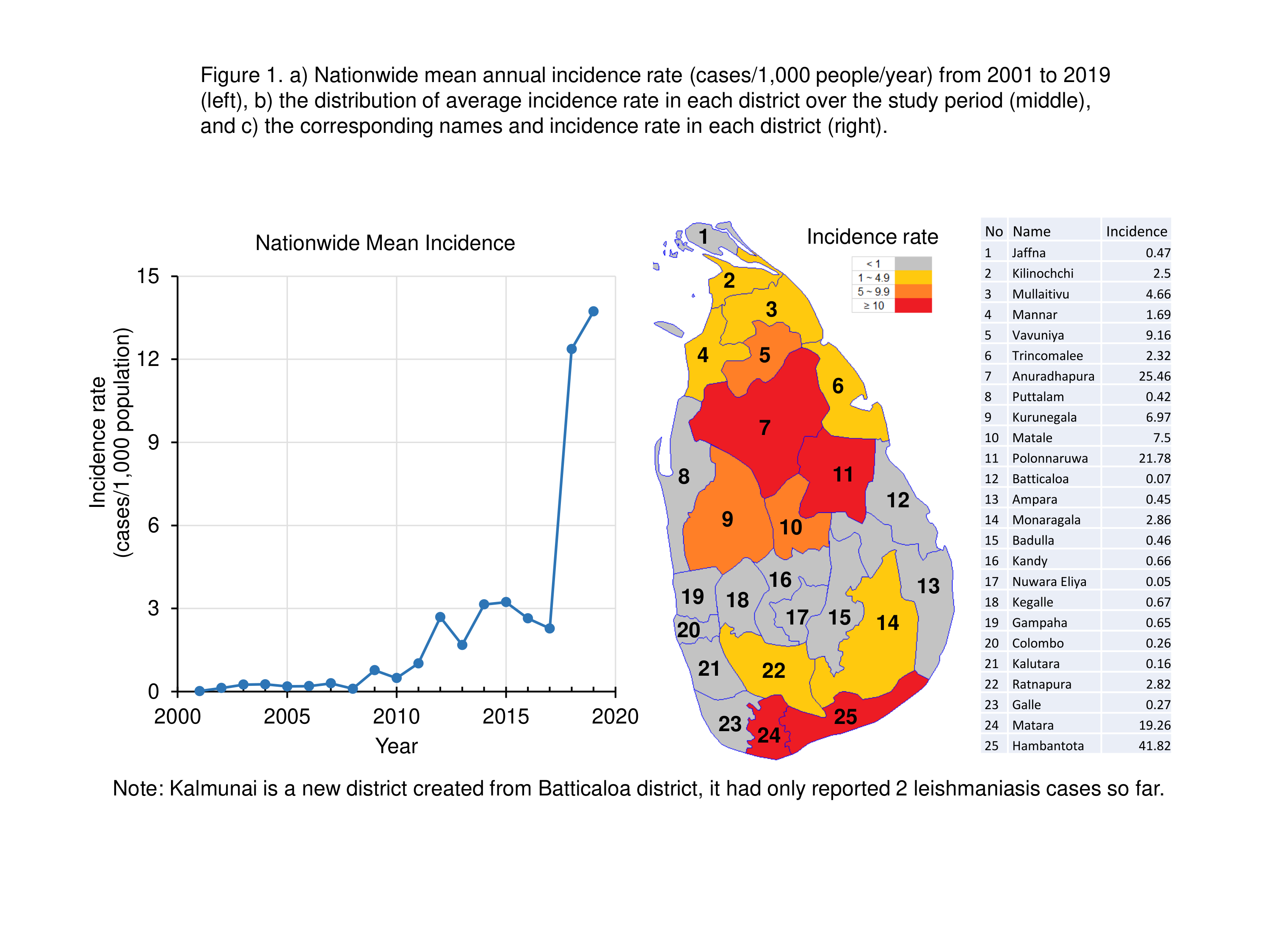}
    \caption{Sri Lanka regional map and mean incidence rate over the last two decades.}
    \label{Sri}
\end{figure*}
\section{Introduction}
Leishmaniases are tropical diseases caused by leishmania parasites and transmitted through the bites of vector sand flies. The 
cutaneous leishmaniasis (CL) is the most common threat and health risk in developing countries in the tropical regions. In this paper, we study data from Sri Lanka that has reported a substantial surge in clinical leishmaniasis cases in the past 20 years (Fig.\ref{Sri}, a)). Previous studies \cite{sri_leish_10,sri_leish_18} found that (1) leishmaniasis epidemics in Sri Lanka had two transmission hot spots, one on the south coast and another in the north central region of the 
country (Fig.\ref{Sri},b)), with a biannual seasonal variation; (2) outdoor activities, including occupational exposure and living near a vector breeding area, are some of the key risk factors of infection.
An important scientific task for public health is to model the spatio-temporal dynamics in leishmaniasis transmission and the driving forces behind it, thereby help
predict future infections and outbreaks. 
\medskip

In this paper, we aim to generalize and advance existing geo-statistical and ecological models \cite{geo_statsrev_99,st_stats07}
by incorporating spatio-temporal  transmission factors such as climate effects and local carryover of infections from neighboring regions. Our main contributions are:
\medskip

\noindent (1) modeling  
leishmaniasis spread between neighboring areas by a recurrent neural network with input data from up to three most infected neighbors;
\medskip

\noindent (2) including climate data input as an external factor, since the development of both the sand flies and the parasites inside their guts are affected by climatic conditions;
\medskip

\noindent (3) hybridizing (1) and (2) with regression to form an integrated nonlinear space-time model trained by stochastic gradient descent on 51 months (2013-03 to 2017-08) and tested on 18 months (2017-09 to 2018-12) in 5 highly infected regions of Sri Lanka. 
\medskip

The rest of the paper is organized as follows. In section 2, we review related prior work on infectious disease modeling where climate and geo-neighbor factors have been separately modeled. In section 3, 
we outline pre-processing of raw data to remove trend, and introduce our integrated model structure with embedding operations of climate and time stamps (monthly)
motivated by design of recommender systems. 
In section 4, we go over training and test data,  and compare prediction results with ARIMA as baseline. In terms of both root mean squares error and maximum absolute error, our integrated model, when applied on the difference input data, out-performed ARIMA significantly in 5 Sri Lanka regions with high infections. Moreover, adding climate data consistently improves 
prediction, which supports climate as a strong correlate to fly and parasite mediated transmissions. 

\begin{figure}[ht!]
\centering
  \subfloat[]{\includegraphics[width=7cm]{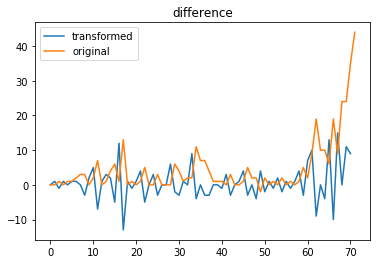}\label{fig2}}
  
  \subfloat[]{\includegraphics[width=7cm]{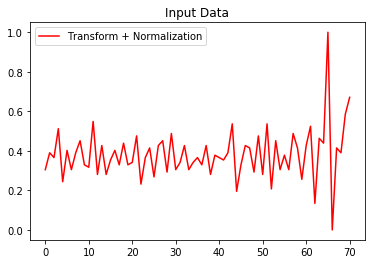}\label{fig1}}
   \caption{An example of data pre-processing for region Kurunegala: a) raw and transformed (differenced) data; (b) transformed (differenced) and normalized data. }\label{fig:transform}
\end{figure}
\section{Related Prior Works}
Forecasting of disease time series 
has gone far beyond the traditional  regression approach. External information has been widely used in models; for example, \cite{argo} proposed Auto-Regression method with GOogle search data (ARGO) that used google search information as additional regressors. The intuition of ARGO is that the amount of searches of influenza and related key words indicate the outbreak of influenza. Utilizing this external internet search information, ARGO outperforms auto-regressive model and its variant ARIMA on CDC influenza data. Unfortunately, Google Correlate, the website where Google provided the internet search data, has been shut down for many years. Motivated by the infectious nature of the influenza, \cite{srnn} proposed graph-structured recurrent neural networks (SRNN) to learn the interaction of geographical spread of influenza. As a result, SRNN further improves ARGO's accuracy on CDC data. Recently, spatio-temporal models combining epidemic differential equations and RNNs \cite{li2020recurrent, zheng2020spatial} have been proposed for one to seven day ahead forecasting of Covid-19 cases in Italy and the US. For vector-borne diseases, due to the difficulty of tracking fly populations and evolution, mixed linear regression-autoregression models with near-neighbor spatial coupling has been commonly used for prediction and risk analysis \cite{geo_statsrev_99,st_stats07}.

\section{Our Approach}
\subsection{Transformation of Raw Data}
Since the original leishmaniasis data is highly non-stationary \cite{sri_climate_06}, we first transform (pre-process) it to be approximately stationary. There are several popular techniques to stationarize data. We found that taking the first order difference along time is effective to improve the stationarity of the raw data here. In Fig. \ref{fig:transform}, panel (a) shows how the original case data of Kurunegala region is transformed by the difference method. As is well-known that RNN is sensitive to normalization, panel (b) shows the normalized-transformed data that will be fed into our model. We remark that normalization is a stationary-invariant process. 

\subsection{Basic Space-Time Model}
In classical space-time geological and ecological statistical modeling (\cite{geo_statsrev_99,st_stats07} among others), the cases of neighbors are summed as a single regressor. To learn the impact of neighbors more at depth, we use RNN to process such information and extract (``edge'') features as in \cite{li2020recurrent,zheng2020spatial}. Let $\mathbf{y}_{e,t}=(y^1_{t-1}, y^2_{t-1}, y^3_{t-1})$ be a vector of observations from the three neighbors that have the highest cases at $t-1$. Define:
$$h_t = LSTM(\mathbf{y}_{e,t})$$
$$f_t = w^Th_t$$
where LSTM is a standard long short-term memory network \cite{lstm}. If a region has less than three neighbors, we pad zero into $\mathbf{y}_e$. Let $I$ be the set of neighbors, then the model output (an estimated case number for a region of interest at time $t$) is: 
\begin{equation}
    \hat{y}_t= \sigma\big(\alpha y_{t-1}+\beta \sum_{i\in I}y^{i}_{t-1}+f_t+b\big),
    \label{eq:eq1}
\end{equation}
where $\sigma := \max(x,0)$ is the rectified linear unit (ReLu) activation function. 

\begin{figure}[ht!]
    \centering
    \includegraphics[scale=0.5]{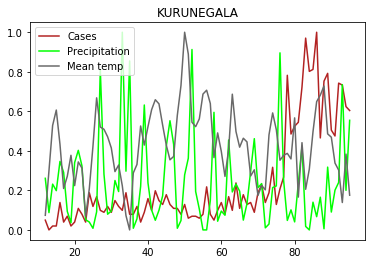}
    \caption{Normalized data of cases vs. mean temperature and precipitation in Kurunegala region of Sri Lanka.}
    \label{fig:temp}
\end{figure}
\begin{figure*}[]
    \centering
    \includegraphics[height = 7cm, width =15cm]{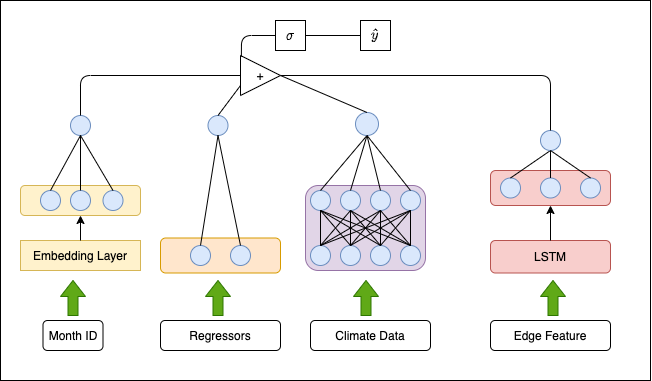}
    \caption{Illustration of our model  architecture (viz. equation (2)).}
    \label{fig:model}
\end{figure*}

\subsection{Integrated Model}
According to previous studies (  \cite{zhou2004association,sri_climate_06} among others), leishmaniasis outbreak is highly correlated to climate conditions such as temperature, rainfalls and seasonality, see Fig. 3 and Fig. 5 for illustrations. Hence, we adopt climate data as an external feature to further improve our model performance. The climate data, denoted as $\mathbf{v}_t\in \mathbb{R}^4$ being part of our model input, contains the maximum temperature, minimum temperature, mean temperature, and precipitation during month $t$. Unfortunately, RNN is not good at directly learning the impact of external feature $\mathbf{v}_t$ based on our experiments. The temperature effect turns out to be highly non-linear. Hence, we set out to learn the order-2 interactions of the climate features. The complete order-2 interactions of $n$ features involve ${n \choose 2}+n$ additional features, which are expensive to compute and can easily lead to over-fitting. Similar problem arises in capturing high-order interactions of user-item features in recommendation systems where the so-called cross layer method \cite{cross1} is proposed through Hadamard product and a weight matrix and the interaction is encoded into a vector of length $n$. As in  \cite{cross1}, we compute the order 2 interaction of climate features as follows:
$$
    (W \mathbf{v})\odot \mathbf{v} = \begin{bmatrix}
    w_{11}v_1^2 + w_{12}v_1v_2 + w_{13}v_1v_3 +w_{14}v_1v_4\\
    w_{21}v_1v_2 + w_{22}v_2^2 + w_{23}v_1v_3 + w_{24}v_2v_4\\
    w_{31}v_1v_3 + w_{32}v_2v_3 + w_{33}v_3^2 + w_{34}v_3v_4\\
    w_{41}v_1v_4 + w_{42}v_2v_4 + w_{43}v_3v_4 + w_{44}v_4^2
    \end{bmatrix}
$$
Once the order interaction of climate features has been encoded in $(W\mathbf{v}\odot\mathbf{v})$, we use a dense layer to map the interaction of climate data into the final prediction:
$$\psi(\mathbf{v})=\Tilde{w}^T[(W\mathbf{v}\odot \mathbf{v})].$$

In addition to external features, capturing seasonality has drawn much attention in recent literature \cite{prophet, zhou2004association} of time-series forecasting. A classical analytical approach is to use partial sums of Fourier series to represent seasonality.
However, the performance relies on fine tuning a non-trainable integer parameter (the number of terms). Instead, in view of  the personalization technique of recommendation system \cite{embed}, we employ an embedding layer to map the month ID, 0 to 11, to a higher dimension
to learn seasonality of the data. Then, we use a dense layer to map it into the output:
$$g(t)= \hat{w}^T(\text{embed(t)}).$$
Integrated with the climate features and seasonality, our model is formulated as:
\begin{equation}
    \hat{y}_t= \sigma\big(\alpha y_{t-1}+\beta \sum_{i\in I}y^{i}_{t-1}+f_t+g(t)+\psi(\mathbf{v_t})+b\big).
    \label{eq:eq2}
\end{equation}
The architecture of our model is illustrated in Fig. \ref{fig:model}. As shown in equation (\ref{eq:eq2}), the final prediction model is written as the sum of all learned information followed by ReLu activation function. The training loss function is 
$$\mathcal{L}(\Theta)=\sum_{t=1}^{N}(\hat{y}_t-y_t)^2$$
which is minimized by an adaptive Adam optimizer to arrive at an optimal value $\Theta^*$. 
\section{Experimental Results}
Our clinically confirmed leishmaniasis case data came from from the national diagnostic and research laboratory at the University of Colombo, Sri Lanka; the epidemiology unit of the Sri Lanka Ministry of Health and through communication with medical health officers.  The climatic data came from meteorological stations in Sri Lanka in the format of maximum/minimum/mean temperature and precipitation.
We use 51 months (2013-03 to 2017-08) for training and 18 months (2017-09 to 2018-12) for testing, with ARIMA as our baseline model. We set the standard parameters of ARIMA as $(p,q,d)=(2,1,1)$ after optimizing. Note that with $d=1$, ARIME also applied difference transform to the original data. Meanwhile, we compare the performance of our base model equation (\ref{eq:eq1}) and the integrated model equation (\ref{eq:eq2}). The results are shown in Table \ref{tab:RMSE} and Table \ref{tab:MAE}. We evaluate the models using both MAE and RMSE metrics. 
Let $\mathbf{e}=\{e_i|e_i = |\hat{y}_t-y_t|, i \leq n\}$, where $n$ is the number of data points in testing set. Then,
$$\text{MAE}=\sum_{i=1}^{n}\frac{|e_i|}{n}$$
$$\text{RMSE}=\sqrt{\sum_{i=1}^{n}\frac{e_i^2}{n}}$$
We note that RMSE $\geq$ MAE, and it can be shown that RMSE$-$MAE $= \text{Var}(\mathbf{e})$.
We observe that equation (\ref{eq:eq1}) outperforms ARIMA model in both MAE and RMSE prediction errors. 

\begin{figure}
    \centering
    \includegraphics[scale=0.5]{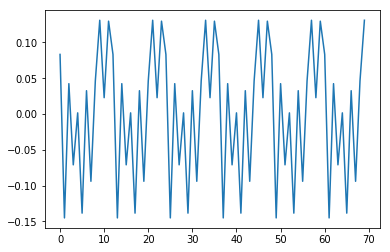}
    \caption{Seasonality component $g(t)$ of Polonnaruwa}
    \label{fig:my_label}
\end{figure}

\begin{table}[ht!]
    \centering
\begin{tabular}{c|c|c|c}
    \hline
        Region   & (1)  & (2) &ARIMA \\
    \hline
     Matara& 11.38& \bf{11.36} & 11.92\\
     Anuradhapura&14.70&\bf{12.81}&14.70\\
     Polonnaruwa & 8.65&\bf{8.04}&8.65\\
     Kurunegala & 14.70 &\bf{10.08} & 10.35\\
     Hambantota & 32.76&\bf{32.12} &34.63 \\
     \hline

\end{tabular}

\caption{RMSE prediction errors of different models}
\label{tab:RMSE}
\end{table}
\begin{table}[ht!]
    \centering
\begin{tabular}{c|c|c|c}
    \hline
       Region        & (1)  & (2) & ARIMA \\
    \hline
     Matara& 9.33& \bf{8.62} & 9.54\\
     Anuradhapura&10.25&\bf{9.88}&12.2\\
     Polonnaruwa & 7.20&\bf{6.08}&7.57\\
     Kurunegala & 17.50 &\bf{16.80} & 17.50 \\
     Hambantota & 41.28& \bf{39.77}& 41.28\\ 
     \hline

\end{tabular}

\caption{MAE errors of different models}
\label{tab:MAE}
\end{table}
Equation (\ref{eq:eq2}) performs the best among the three models. Hence, the edge features in cases of neighboring regions helped model (1) to outperform ARIMA, which is only based on historical observations of the region of interest. The external climate (see Fig. \ref{fig:my_label} for a  seasonality illustration) information helps model (2) to further improve  prediction. 

\section{Conclusion}
In this study, we integrated components of geographical spatial  information, temperature, and seasonality to build a spatio-temporal network model for predicting vector-borne disease cases. We employ the cross layer from recommendation system to compute the order-2 interaction of climate data, and utilize embedding layer to map month ID to higher dimensions to learn seasonality. 
The model is successfully trained on leishmaniasis data of several regions in Sri Lanka with high infections 
(see Fig. \ref{fig:model}). 

In future work, we plan to study other vector-borne disease data with our model (2), and also generalize RNN to an efficient transformer model to explore additional non-local temporal information for improving prediction. 

As suggested in Fig. 3, the effects of climatic data may have a latent period
to induce vector growth and subsequent case upswing. In future work, we plan to introduce a time delay in the climate term of our model and learn it from the data for another improvement.   

\section{Acknowledgements}
This work was partially supported by NSF grants DMS-1924548 and DMS-1952644.
\begin{figure*}[]
    \centering
    \includegraphics[height = 10cm, width =12cm]{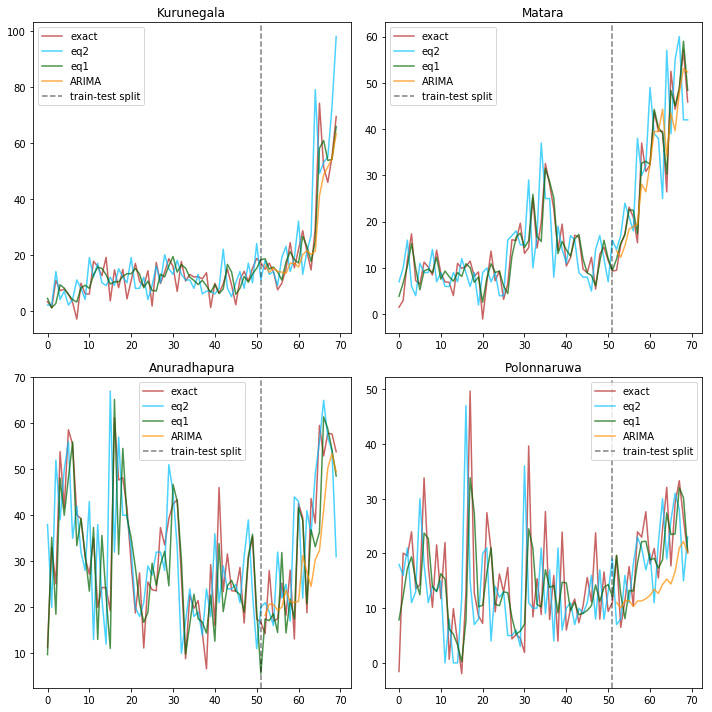}
    \caption{Training and prediction (separated by dashed line) of models (1) and (2) vs. ARIMA in 4 regions of Sri Lanka.}
    \label{fig:model}
\end{figure*}
\bibliography{example}
\end{document}